\documentclass[journal]{IEEEtran}

\ifCLASSINFOpdf
\else
\fi

\usepackage{color}
\usepackage{bm}
\usepackage[ruled,vlined]{algorithm2e}
\usepackage{tikz} 
\usepackage{amssymb}
\usepackage{cite}
\usepackage{graphicx}
\usepackage{epsfig,subfigure}
\usepackage{amsmath}
\usepackage{algorithmic}
\usepackage{fancyhdr}
\usepackage{caption}
\usepackage{hyperref}
\usepackage{authblk}

\begin{document}
\title{Improving Disentangled Representation Learning\\ 
with the Beta Bernoulli Process}

\author[1]{Prashnna Kumar Gyawali}
\author[1]{Zhiyuan Li}
\author[1]{Cameron Knight}
\author[1]{Sandesh Ghimire}
\author[2]{B Milan Hora\`cek}
\author[2]{John Sapp}
\author[1]{Linwei Wang}
\affil[1]{Rochester Institute of Technology, NY, USA}
\affil[2]{Dalhousie University, Halifax, Canada}



\maketitle

\begin{abstract}
To improve the ability of variational auto-encoders (VAE) to disentangle in the latent space,
existing works mostly focus on enforcing the independence among the learned latent factors.  
However, 
the ability of these models to disentangle often decreases as the complexity of the generative factors increases. 
In this paper, 
we investigate the little-explored effect of the modeling capacity of a posterior density on the disentangling ability of the VAE. 
We note that the independence within and the complexity of the latent density are two different properties we constrain 
when regularizing the posterior density: 
while the former promotes the disentangling ability  of  VAE,  the  latter --  if  overly  limited --  creates  an unnecessary competition with the data reconstruction objective in VAE.  
Therefore,  if  we  preserve the  independence  but  allow  richer  modeling  capacity  in  the posterior  density,  we  will  lift  this  competition  and  thereby allow  improved  independence  and  data  reconstruction  at  the same  time. 
We investigate this theoretical intuition with a VAE that utilizes a non-parametric latent factor model, the Indian Buffet Process (IBP), as a latent density 
that is able to grow with the complexity of the data. 
Across three widely-used benchmark data sets (MNIST, 3D Chairs and dSprites) 
and two clinical data sets little explored for disentangled learning, 
  we qualitatively and \textit{quantitatively} demonstrated the improved disentangling performance of IBP-VAE over the  state of the art. In the latter two clinical data sets riddled with complex factors of variations, we further demonstrated that unsupervised disentangling of nuisance factors via 
 IBP-VAE -- when combined with a supervised objective -- can not only improve task accuracy in 
 comparison to relevant supervised deep architectures, 
but also facilitate knowledge discovery related to task decision-making. 
A shorter version of this work will appear in the ICDM 2019 conference proceedings.
\end{abstract}


\IEEEpeerreviewmaketitle

\section{Introduction}
\IEEEPARstart{A}{n} inherent goal in deep learning is to distill   task-relevant latent representations 
that are  
invariant to other  nuisance
factors 
in the data. 
State-of-the-art 
deep neural networks 
achieve this by  
careful engineering of the network architecture,  
along with supervised training 
with a large number of task labels. 
However, 
the effectiveness of supervised training
relies heavily on
data quantity and label quality, 
especially in data with  
a wide range of 
data-specific 
factors of variations.  
Moreover, 
interpreting the results 
of these networks
-- important in areas such as 
clinical tasks -- 
remains 
challenging 
\cite{chakraborty2017interpretability}.



Unsupervised disentangled representation learning 
provides a task-agnostic approach 
to learn latent generative factors 
that are semantically interpretable 
and mutually invariant
\cite{higgins2016beta,chen2018isolating,kim2018disentangling,kumar2017variational,chen2016infogan}.  
Many of recent successes in this area 
are based on variational autoencoders (VAE), 
which modernize variational inference by using neural networks to 
parameterize both 
the likelihood of data $\mathbf{x}$ given 
latent variable $\mathbf{y}$, $p_{\bm{\theta}}(\mathbf{x}|\mathbf{y})$, 
and the approximated posterior density of $\mathbf{y}$, $q_{\bm{\phi}}(\mathbf{y}|\mathbf{x})$ \cite{kingma2013auto, DBLP:conf/icml/RezendeMW14}. 
The objective of VAE training is thus 
to maximize the variational evidence lower bound (ELBO)
of the marginal data likelihood: 
\begin{equation}
  \label{eq:VAEObj}
  \log{p}(\mathbf{x}) \geq \mathcal{L}  = \mathop{\mathbb{E}_{q_{\bm{\phi}}(\mathbf{y}|\mathbf{x})}}[\log{p_{\bm{\theta}}(\mathbf{x}|\mathbf{y})}] - KL(q_{\bm{\phi}}(\mathbf{y}|\mathbf{x})||p(\mathbf{y}))
\end{equation} 
where the first term can be 
interpreted as  
data reconstruction, 
while the second penalty term 
constrains the approximated posterior density $q_{\bm{\phi}}(\mathbf{y}|\mathbf{x})$ 
to be similar to a prior $p(\mathbf{y})$ 
by minimizing their 
Kullback-Leibler (KL) divergence. 

To improve disentangled learning in VAE, 
the primary focus 
has been 
on enforcing independence 
among the learned latent factors,  
achieved by more heavily penalizing  
the distance from
$q_{\bm{\phi}}(\mathbf{y}|\mathbf{x})$ 
\cite{higgins2016beta} 
or 
its marginal density $q_{\bm{\phi}}(\mathbf{y})$ \cite{kumar2017variational} to a prior $p(\mathbf{y})$ that is independent among dimensions. 
This may be strengthened by an 
explicit 
independence penalty on
$q_{\bm{\phi}}(\mathbf{y})$, 
\emph{e.g.}, either  
added to the ELBO \cite{kim2018disentangling} or 
isolated from the ELBO 
through total-correlation decomposition \cite{chen2018isolating}. 
These investigations, however, 
are carried out in the context of a Gaussian approximation of the posterior density, 
limiting its ability 
to model generative factors 
with increased complexity \cite{higgins2016beta}.  

In parallel, 
enabling richer posterior approximations 
has been an active topic of interest 
for improving 
data reconstructions 
in VAE
\cite{tomczak2017vae,kingma2016improved}. 
This is often 
achieved by designing more complex densities 
$q_{\bm{\phi}}(\mathbf{y}|\mathbf{x})$ and/or 
$p(\mathbf{y})$ 
with increased modeling power. 
Their effect on the disentangling ability of the VAE, however, has not been considered.

In this work, we investigate 
the little-explored 
relationship between the modeling capacity of a posterior density and the disentangling ability of the VAE. 
Following \cite{chen2018isolating}, 
we note that when constraining 
a (marginal) posterior density 
to an independent prior, 
we enforce two effects: 
the independence among the latent factors, 
and the complexity of the density. The former has to do with disentangling, while the latter affects the modeling capacity of VAE: 
when enforcing an independent density with limited modeling capacity, 
the latter creates an unnecessary tension 
with the reconstructing objective (data likelihood). 
Therefore,  
alternative to directly reinforcing the independence, 
we rationalize that a richer modeling capacity   
will indirectly improve disentangling 
by reducing this tension. 

Formally, we hypothesize 
that \textit{an independent latent factor model with  
increased modeling capacity} 
will improve disentangled learning 
of generative factors 
with increased complexity. 
We investigate this theoretical intuition 
with a VAE model 
that utilizes a 
non-parametric Bayesian latent factor model -- the Beta Bernoulli process implemented via the Indian Buffet process (IBP) \cite{griffiths2011indian} -- to model an unbounded number of 
mutually independent 
latent factors. 
We first evaluate this IBP-VAE 
on three 
benchmark data sets 
(color-augmented MNIST \cite{lecun1998gradient}, 3D Chairs \cite{aubry2014seeing}, and dSprites 
\cite{dsprites17}), 
where we \emph{qualitatively and quantitatively} 
demonstrate 
its improved ability 
to disentangle a 
variety of 
discrete and continuous 
generative factors 
in comparison to 
state of the arts 
\cite{higgins2016beta,chen2018isolating}.  
Furthermore, 
supporting our theoretical intuition, 
we show that IBP-VAE was able 
1) to achieve 
improved data reconstruction as well as improved independence within the learned posterior densities compared to the use of an independent Gaussian density, and 
2) to achieve better disentanglement compared to the use of a complex density that does not consider independence. 

We then 
further demonstrate 
in two distinct clinical data sets
that -- when combined with task labels -- unsupervised learning 
of nuisance factors 
can help improve the 
extraction of task-relevant representations 
while facilitating 
the discovery of knowledge 
related to task decision-making.  
We considered a 
skin lesion image data set \cite{gutman2016skin} 
where the primary task is to 
classify malignant skin lesion (melanoma) 
from benign lesion, 
challenged by 
the need to extract subtle features relevant to melanoma detection 
(\textit{e.g.}, color and shape asymmetry) from a large variety of lesion features 
\cite{argenziano1998epiluminescence}. 
We also considered
a 
clinical electrocardiogram (ECG) data set \cite{sapp2017real} where the primary task is 
to localize the origin of arrhythmia beat in the heart  
from the morphology of 12-lead ECG signals, 
challenged by an unknown number of 
nuisance factors including 
patient demographics, 
geometry,  
and pathology 
that
affect ECG morphology through complex physiological processes. 
These challenges 
were evident from 
the limited performance of relevant 
supervised deep architectures on each data set, 
which we show could be improved by adding unsupervised disentangling of nuisance factors via IBP-VAE. 
Note that 
the effectiveness of 
disentangled representation learning, 
either fully unsupervised 
or combined with supervised tasks, 
has been little investigated 
in this type of clinical data sets. 

To summarize, the main contributions of this work include:
\vspace{-.4cm}
\begin{itemize}
    \item Departing from current focus on the independence of latent factors for improving disentangled representation learning, we theoretically rationalize that a richer posterior approximation, with preserved independence, will improve disentangling of generative factors 
    by indirectly reducing the tension between the disentangling and reconstructing capacity of VAE.
    \item Via an IBP-VAE with an infinite latent factor posterior approximation, 
    we qualitatively and quantitatively verify our hypothesis 
    on widely-used benchmark data sets. 
    \item We further demonstrate -- for the first time on clinical data sets little explored for disentangled representation learning -- that unsupervised disentangling of nuisance factors  will improve supervised tasks and facilitate discovery of semantic factors relevant to task decision-making.
\end{itemize}
Overall,  
while significant progresses of 
disentangled representation learning 
have been demonstrated 
on visual benchmarks with 
relatively well-known 
generative factors, 
its feasibility is little known
in real-world data sets --- 
such as clinical images and signals ---  
where there is a large and often unknown number of generative factors with a complex relationship with the data.
We hope this work will contribute to bringing 
unsupervised 
disentanglement learning towards this direction. 

\section{Related Works}
Recent developments of 
unsupervised disentangled representation learning 
are primarily considered in the context of deep generative models, 
such as VAE \cite{kingma2013auto,DBLP:conf/icml/RezendeMW14} and 
generative adversarial networks 
(GAN) \cite{goodfellow2014generative}. 
In $\beta$-VAE \cite{higgins2016beta}, 
it was demonstrated that 
unsupervised disentanglement can be achieved by
constraining the posterior density of the latent representation to be similar to an isotropic Gaussian prior 
with independence among each latent dimension. 
Following this line of rationale,
better enforcing the independence 
among latent dimensions has been 
a main approach to improving disentangled learning in VAE. 
Examples include 
adding to the ELBO a penalty  constraining the marginal posterior density $q_{\bm{\phi}}(\mathbf{y})$ to be similar to an independent prior  
\cite{kumar2017variational}, 
or directly penalizing the dependence within 
$q_{\bm{\phi}}(\mathbf{y})$
through a total-correlation term, $KL(q_{\bm{\phi}}(\mathbf{y})||\prod_i q_{\bm{\phi}}(y_i))$, either isolated from the ELBO
\cite{chen2018isolating} 
or added to the ELBO objective 
\cite{kim2018disentangling}.  
In the context of GAN, 
it was also shown that 
maximizing the mutual information 
between the latent representation 
and data 
can help 
learning 
disentangled 
representations 
\cite{chen2016infogan}. 
These disentangling-focused networks, 
however, 
do not consider 
the modeling capacity of the latent densities: 
on the contrary, 
using a common choice of 
independent Gaussian densities, 
the disentangling ability of these networks 
generally decreases 
as the number of generative factors in the data 
increases \cite{higgins2016beta}. 

In parallel, 
it has been widely discussed 
that a Gaussian assumption for the posterior density
may underestimate 
the required complexity of the marginal posterior of 
the latent representation \cite{hoffman2016elbo, alemi2018fixing,burda2015importance}.
There has been an increased interest in 
enabling richer posterior approximations
in VAE, 
including means to accommodate 
Gaussian mixture models \cite{dilokthanakul2016deep},  
autoregressive models \cite{chen2016variational}, flow-based models \cite{kingma2016improved} 
and 
Bayesian nonparametric 
models in VAE
\cite{nalisnick2016stick,goyal2017nonparametric,chatzis2014indian,singhstructured}.
In specific, 
non-parametric IBP has been previously considered 
in VAE
\cite{chatzis2014indian,singhstructured}. 
However, 
while demonstrating an improvement 
in data reconstruction and 
posterior approximations, 
these works do not consider 
the role of richer posterior densities in 
learning disentangled representations.

The presented work can be seen as an attempt to bridge  the above two lines of works. We theoretically rationalize that the independence within and the modeling capacity of the latent density are two separate effects when we regularize the posterior density: the former affects disentangling, while the latter affects data reconstruction. 
Alternative to 
directly manipulating independence,  
we bring a new perspective that 
richer posterior approximations, 
with preserved independence, 
will indirectly 
facilitate disentangling by reducing its competition with the reconstruction objective in ELBO.
This is to our knowledge the first investigation of the role of 
posterior modeling capacity in disentangled representation learning.

Regarding the separation of nuisance factors for learning task-related representations,
this work is marginally related to \textit{fair} representation
learning  \cite{zemel2013learning,louizos2015variational} 
and its applications in confounder filtering \cite{wu2018fair}.
The notion of learning \textit{fair} representations was introduced in \cite{zemel2013learning} and later extended to VAE in  \cite{louizos2015variational} to  obfuscate observed confounding attributes, such as age or sex \cite{zemel2013learning}, from the learned representation. 
In application to health-care domain, an approach was presented in \cite{wu2018fair} to remove the effect of confounding factors by identifying network weights that are associated with confounding factors in the pre-trained model. 
All of these works, however, 
focus on removing a small number of observed nuisance factors 
while the presented work considers 
disentangling an unknown number of 
unobserved nuisance factors. 

\section{Methodology}

\subsection{Preliminaries: Beta-Bernoulli Process}
The Beta-Bernoulli process is a stochastic process that 
defines a probability distribution over a sparse binary matrix indicating feature activation for $K$ features. The generative Beta-Bernoulli process taking the limit $K \rightarrow \infty$ is also referred to as the IBP \cite{griffiths2011indian}. The infinite binary sparse matrix $\mathbf{Z} \in \{0,1\}^{N\times K^+}$ represents latent feature allocation, where $z_{n,k}$ is 1 if feature $k$ is active for the $n^{th}$ sample and 0 otherwise. For practical implementations, stick-breaking construction \cite{teh2007stick} is considered where the samples are drawn as:
\begin{equation}
\begin{aligned}
  \label{eq:independence}
& \nu \sim Beta(\alpha, 1); \pi_k = \prod_{i=1}^{k} \nu_i \\
& z_{n,k}|\pi_k \sim Bernoulli(\pi_{k}) 
\end{aligned}
\end{equation}
where the hyperparameter  $\alpha$ 
represents the expected number of features 
in the data. 

\subsection{Theoretical Intuition}
\label{sec:model}

As introduced earlier, the ELBO objective (\ref{eq:VAEObj}) 
consists of data reconstruction regularized by some constraints on the posterior density. Independence of the posterior density has been one constraint  shown to be effective in improving disentangling \cite{higgins2016beta,chen2017disentangling,kim2018disentangling}. 
To examine the role of other properties of the posterior pdf in disentangling, we delve further into ELBO following the 
 decomposition
 in \cite{hoffman2016elbo,chen2018isolating}: 
\begin{equation}
    \begin{aligned}
    \mathcal{L} = & \underbrace{\frac{1}{N}\sum_{n=1}^{N}\mathbb{E}_{q_{\bm{\phi}}(\mathbf{y}_{n}|\mathbf{x}_{n})}[\log{p_{\bm{\theta}}(\mathbf{x}_{n}|\mathbf{y}_{n})}]}_{\text{Average reconstruction}} \\
    & - \underbrace{KL(q(\mathbf{y})||\prod_{j}q(\mathbf{y}_j))}_{\text{Total Correlation}} - \underbrace{\sum_{j}KL(q(\mathbf{y}_j)||p(\mathbf{y}_j))}_{\text{Dimension-wise KL}} \\
    & - \underbrace{(\log{N} - \mathbb{E}_{q(\mathbf{y})}[\mathbb{H}[q(\mathbf{n}|\mathbf{y})]]))}_{\text{Index-code mutual information}}
    \end{aligned}
    \label{eq:tc-decompo}
\end{equation}

As shown, 
when minimizing the KL-divergence between the posterior and an independent prior density in (\ref{eq:VAEObj}), 
two constraints take effect: we not only promote the independence within 
$q(\mathbf{y})$ (total-correlation term in (\ref{eq:tc-decompo})), but also constraining the shape and complexity of $q(\mathbf{y})$ (the 3rd and 4th term in (\ref{eq:tc-decompo})). 
While the former promotes the disentangling ability of VAE, 
the latter -- if overly limited -- 
creates an unnecessary 
competition with the data reconstruction objective in ELBO (the 1st term in (\ref{eq:tc-decompo})). 
Therefore, 
if we preserve the independence but allow richer modeling capacity in the posterior density, 
we will lift this competition and thereby allow improved independence and data reconstruction at the same time. 
This is the theoretical basis of the presented hypothesis that \textit{an independent latent factor model  with  increased  modeling  capacity} will  improve  disentangled representation learning in VAE. Below, we investigate this hypothesis with an IBP-VAE where the complexity of the posterior density is able to grow with the 
complexity of the data.

\subsection{Disentangling IBP-VAE}
\label{sec:ibp-vae-main}
\subsubsection{Generative model}
We assume that data  $\mathbf{X} =  \{\mathbf{x}_{n}\}_{n=1}^{N}$ is generated by 
latent representations $\mathbf{Y} = \{\mathbf{y}_{n}\}_{n=1}^{N}$  
that follows a non-parametric IBP prior: 
\begin{equation}
\begin{aligned}
  \label{eq:genModel}
  \mathbf{Y} &= \mathbf{Z} \odot \mathbf{A}\\ 
  p(\mathbf{A}) &= \prod_{n=1}^N \mathcal{N}(0, \mathbf{I}_{K^{+}});\\
  p(\mathbf{Z}|\nu) &= 
  \prod_{k=1}^{K \rightarrow \infty}\prod_{n=1}^N Bernoulli(\pi_k),
    \pi_k = \prod_{i=1}^{k}\nu_i\\
p(\nu) &=  Beta(\alpha, 1)
\end{aligned}
\end{equation}
where 
$\mathbf{Z} = \{\mathbf{z}_{n}\}_{n=1}^{N}, \mathbf{A} = \{\mathbf{a}_{n}\}_{n=1}^{N}$, 
$\odot$ is element-wise product, 
and $N$ is the number of data samples. 
This representation 
essentially allows the model 
to infer which latent features captured by $a_{n, k}, k \in \{1, .. K \rightarrow \infty\}$ is active 
for the observed data $\mathbf{x}_{n}$. 
As the active factors for each data point 
are inferred and not fixed, 
this non-parametric model 
is able to grow with the 
complexity of the data. 

As defined in (\ref{eq:genModel}), 
the IBP 
assumes that each data point possesses feature $k$ with independently-generated probability $\pi_{k}$.  
Each $\mathbf{z}_{n}$ is 
also modeled as a product 
of $K \rightarrow \infty$ 
mutually independent Bernoulli distributions. 
Furthermore, 
each $\mathbf{a}_{n}$ 
is also modeled 
with independent dimensions 
via an isotropic Gaussian density.  
The latent representation $\mathbf{Y}$, as an element-wise product 
between $\mathbf{Z}$ and $\mathbf{A}$, 
is therefore also independent among each feature dimension. 
This provides 
a latent factor model 
that is independent among dimensions but 
with a high modeling capacity. 
We then model the likelihood  $p_{\bm{\theta}}(\mathbf{
X}|\mathbf{Z} \odot \mathbf{A})$ to be Gaussian (real-valued observations) or Bernoulli (binary observations) parameterized 
by neural  networks. 

\subsubsection{Inference model}
\begin{figure}[t]
\begin{center}
  \includegraphics[scale=0.40]{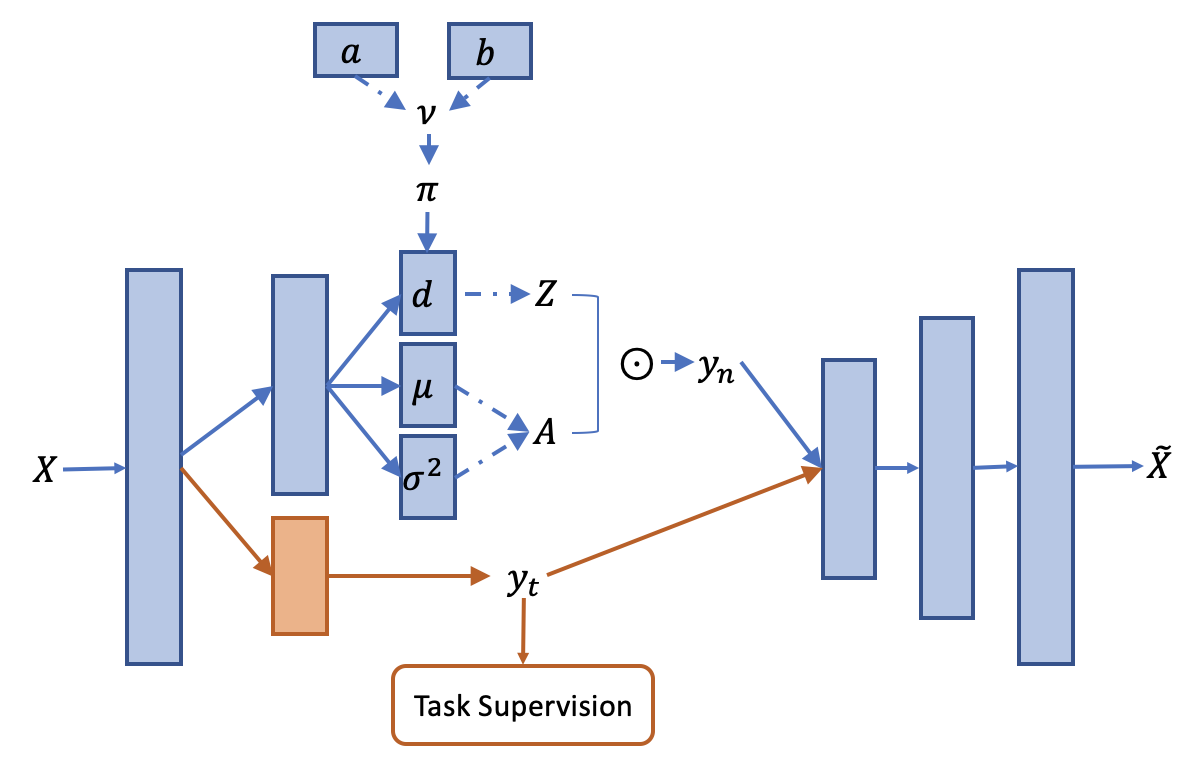}
  \end{center}
  \vspace{-.2cm}
\caption{\small{Outline of the presented IBP-VAE (blue) for unsupervised disentangled representation learning, and cIBP-VAE (blue and orange) for the combination with supervised task learning.}}
\label{fig:GenModel_.}
\end{figure}

We introduce a variational approximation of the posterior density  $q_{\bm{\phi}}(\mathbf{Z},\mathbf{A},\nu |\mathbf{X},a,b)$: 
\begin{equation}
\begin{aligned}
  \label{eq:qDecomposition}
  q_{\bm{\phi}}(\mathbf{Z},\mathbf{A},\nu |\mathbf{X},a,b) &= 
   q(\mathbf{A}|\mathbf{X}) 
  q(\mathbf{Z}|\nu, \mathbf{X}) q(\nu|a,b)
 \\ 
  q(\mathbf{A}|\mathbf{X}) &=\prod_{n=1}^{N} \mathcal{N}(\bm{\mu}(\mathbf{x}_{n}), diag(\bm{\sigma^{2}}(\mathbf{x}_{n})))\\
  q(\mathbf{Z}|\nu, \mathbf{X}) &= \prod_{n=1}^{N} Concrete(\bm{\pi}, \mathbf{d}(\mathbf{x}_n)),
  \pi_k = \prod_{i=1}^{k} \nu_{i}\\
 q(\nu | a, b)
 &= Kumaraswamy(a, b)
  \end{aligned}
\end{equation}
where we use the Concrete distribution  \cite{jang2016categorical, maddison2016concrete} to approximate the Bernoulli distribution, and use the Kumaraswamy distribution \cite{nalisnick2016stick} 
to approximate the Beta distribution (more  
details 
in the Appendix \ref{sup:concrete} and Appendix \ref{sup:kumaraswamy}). 
$q(\nu | a, b)$ are parameterized by $a$ and $b$, and $\mathbf{d}(\mathbf{x}_{n})$, $\bm{\mu}(\mathbf{x}_{n})$ and $\bm{\sigma}^{2}(\mathbf{x}_{n})$ are parameterized by neural networks.
This gives rise to the presented IBP-VAE architecture as illustrated in Fig. \ref{fig:GenModel_.}. 

\subsubsection{Variational inference}
We derive the ELBO, obtained by minimizing the KL divergence between the true posterior and the approximated posterior, for IBP-VAE as:
\begin{eqnarray}
\begin{aligned}
\label{eq:ELBO}
&\log{p}(\mathbf{X}) \geq \mathcal{L}  = \mathop{\mathbb{E}_{q}}[\log\frac{p(\mathbf{X}, \mathbf{Z},\nu, \mathbf{A})}{q(\mathbf{Z},\nu, \mathbf{A}|\mathbf{X},a,b)}] \\
&=\mathop{\mathbb{E}_{q}}\bigg[\log\frac{p(\mathbf{X}|\mathbf{Z},\mathbf{A})p(\mathbf{A})p(\mathbf{Z}|\nu)p(\nu)} {q (\mathbf{A}|\mathbf{X})q(\mathbf{Z}|\mathbf{X},\nu)q(\nu|a,b)}\bigg] \\
& =  \mathbb{E}_{q}[\log{p}(\mathbf{X}|\mathbf{Z}, \mathbf{A})] - KL(q(\nu|a,b)||p(\nu))   \\
& - KL(q(\mathbf{Z}|\nu, \mathbf{X})||p(\mathbf{Z}|\nu)) -KL(q(\mathbf{A}| \mathbf{X})||p(\mathbf{A}))
\end{aligned}
\end{eqnarray}
where $\mathcal{L}$ is optimized with respect to the network weights as well as parameters $a$ and $b$.
This objective function can be interpreted as minimizing a reconstruction error  
along with minimizing the KL divergence between the variational posteriors and the corresponding priors in the remaining terms. 

\subsection{Learning task representations}
We further consider the use of 
disentangled representation learning in supervised learning of task with labels. 
As illustrated in Fig.~\ref{fig:GenModel_.}, 
we split the latent representation $\mathbf{y}$ of data $\mathbf{x}$ into $\mathbf{y}_{ns}$ and $\mathbf{y}_{ts}$. The former represents the nuisance factors 
that will be modeled with the IBP density and learned in an unsupervised manner, 
while the latter is the task-related representation that will be supervised with the task label. 
The likelihood function is now expressed as $p_{\bm{\theta}}(\mathbf{x}|\mathbf{y}_{ns}, \mathbf{y}_{ts})$ and as before is parameterized by the decoder network. 
We encode the nuisance factors $\mathbf{y}_{ns}$ through the stochastic encoder as described earlier, 
and the 
task-representation $\mathbf{y}_{ts}$ with a deterministic encoder 
parameterized by $\bm{\phi}_{ts}$. 
We utilize the task label  by extending the unsupervised objective $\mathcal{L}$ in equation (\ref{eq:ELBO}) with a supervised classification loss on the task representation:
\begin{equation}
\begin{aligned}
  \label{eq:finalObj}
\mathcal{L}^{\gamma} =  \mathcal{L} + \zeta \cdot \mathbb{E}_{p(\mathbf{x},\mathbf{y}_{ts})} [-\log{q_{\bm{\phi}_{ts}}(\mathbf{y}_{ts}|\mathbf{x})}]
\end{aligned}
\end{equation}
where the hyper-parameter $\zeta$ controls the relative weight between the generative and discriminative learning, and $q_{\bm{\phi}_{ts}}(\mathbf{y}_{ts}|x)$ is the label predictive distribution \cite{kingma2014semi} approximated by the deterministic encoder.  
We refer this extension as cIBP-VAE throughout this paper.

\begin{figure}[t]
\begin{center}
\includegraphics[height=180 pt,width=250 pt]{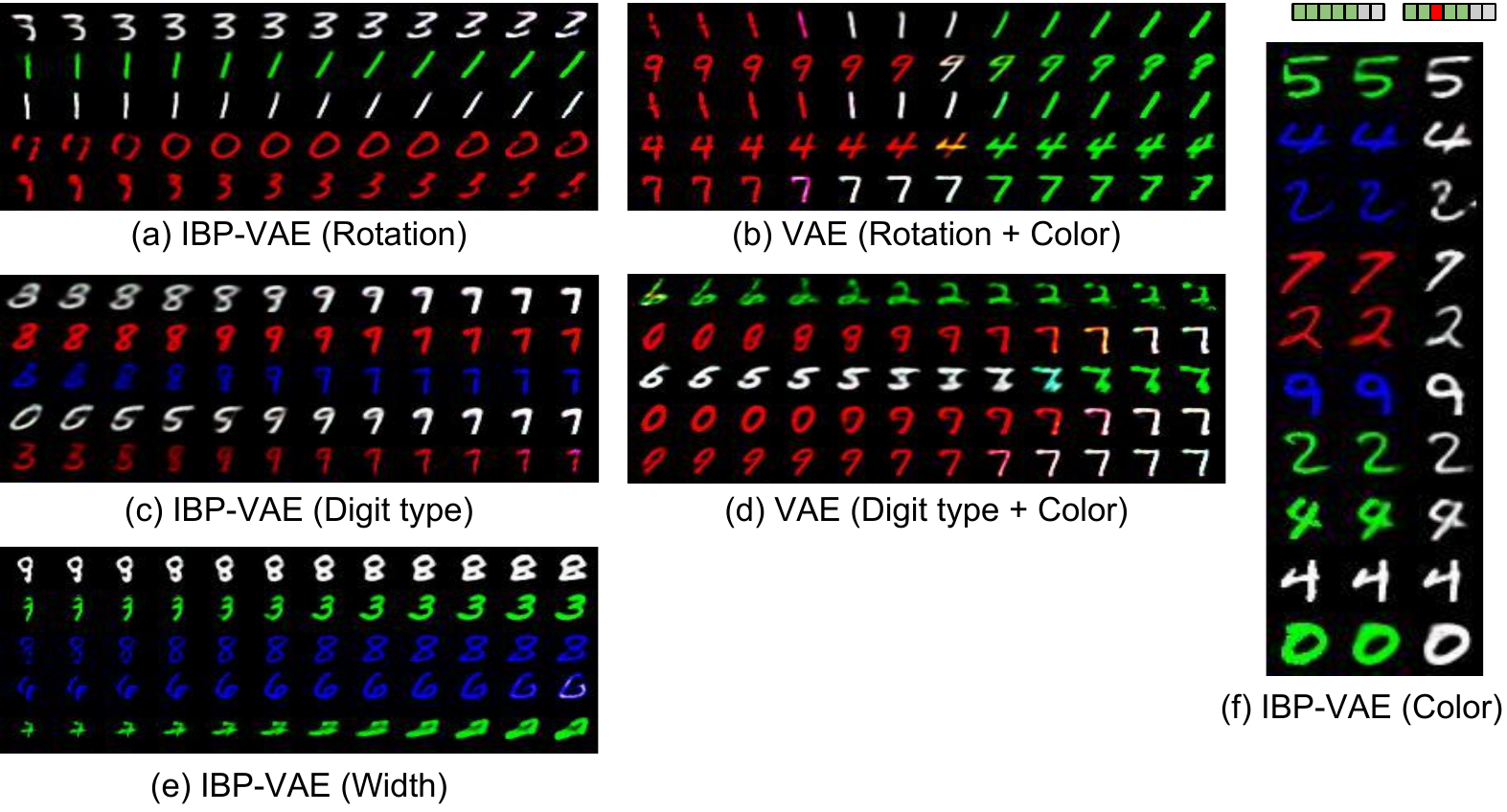}
\end{center}
\vspace{-.4cm}
\caption{\small{[Best viewed in color] (a)-(e): Images generated by traversal along a single latent unit (over a range of [-3, 3]) on the latent representation encoded from a random sample (each row).  
(f): \textit{Triggering} capacity of the IBP-VAE: column one: original images; column two: reconstructed images;  column three: reconstructed images after deactivating the \textit{triggering} unit. The schematic boxes illustrate active (green), de-activated (red), and inactive (grey) units of $\mathbf{z}_{n}$.}}
\vspace{-.2cm}
\label{fig:MNIST_disentanglement}
\end{figure}

\begin{figure}[t]
\begin{center}
\includegraphics[height=320 pt, width=250 pt]{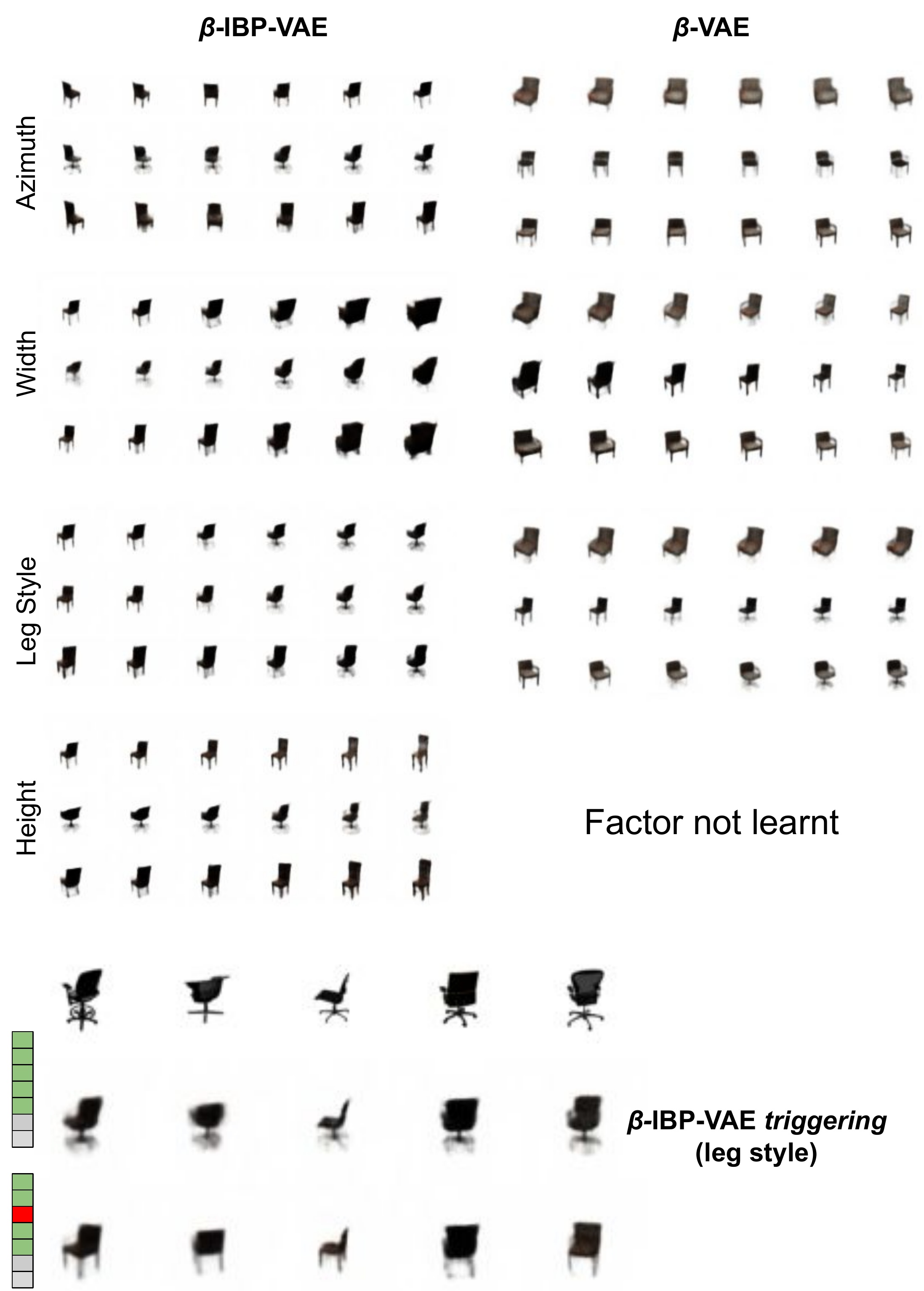}
\end{center}
\vspace{-.3cm}
\caption{\small{(Top) Learned latent variables using $\beta$-VAE and $\beta$-IBP-VAE for the traversal range of (-1, 1). (Bottom) Triggering capacity of the $\beta$-IBP-VAE where the three rows give examples of the original images, reconstructed images, and reconstructed images with the triggering unit for leg styles de-activated.}
\vspace{-.4cm}
}
\label{fig:3D_chair}
\end{figure}


\begin{figure*}[t]
\begin{center}
    \includegraphics[width=\linewidth]{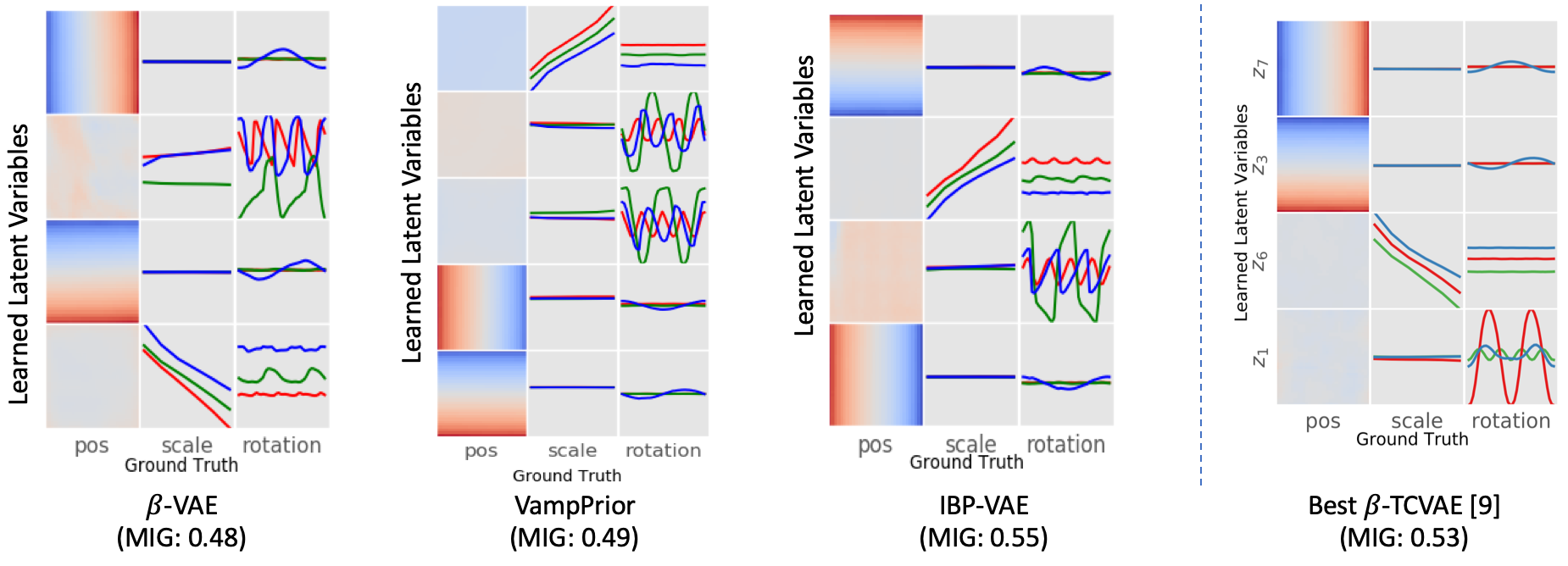}
  \end{center}
  \vspace{-0.4cm}
\caption{\small{[Best viewed in color] Disentangling performance of $\beta$-IBP-VAE compared with $\beta$-VAE and $\beta$-VampPrior at $\beta$ = 5. The best result reported in literature (rightmost plot, reprinted from $\beta$-TCVAE \cite{chen2018isolating} with permission) is also presented for comparison. 
Each plot shows the relationship between each learned latent dimension (row) and each ground-truth factor (column): color in column one encodes high to low values from blue to red; colored lines in column two and three represent 
different object shapes. 
The MIG score (the higher the better) is given for each model. 
}}
\vspace{-0.4cm}
\label{fig:dSprites}
\end{figure*}

\section{Experiments}
\label{experiments}
We performed three sets of experiments in five distinct data sets. This includes three widely-used benchmark data sets for 
unsupervised disentangled representation learning, 
and two real-world clinical data sets 
with their respective tasks of interest. 
Across all data sets, 
we evaluated the disentangling performance of the presented IBP-VAE 
in comparison to 
VAE using a standard isotropic Gaussian prior, varying the regularization parameter $\beta$ for the KL penalty term in both settings (\textit{i.e.}, 
similar to $\beta$-VAE \cite{higgins2016beta}, we use the term $\beta$-IBP-VAE when $\beta>1$ is used with IBP-VAE). 
In the quantitative analysis of disentanglement, we further included comparisons to VAE that uses a complex prior in the form of VampPrior \cite{tomczak2017vae}.  
In the two clinical data sets, 
we also evaluated the performance of  
cIBP-VAE in the respective clinical task 
in comparison to 
supervised 
deep networks 
as well as c-VAE (similar to cIBP-VAE except the nuisance factor follows a Gaussian prior).

Given the diversity of data sets being considered, we leave data and implementation details 
to each subsection. In all experiments, 
implementations of VAE (or c-VAE) adopted the same parameters and architecture as IBP-VAE (or cIBP-VAE) whenever possible.
 Architecture details of other baseline models are included in the Appendix \ref{supp:experiments}. 
In all experiments except dSprites, a validation set was used to select the hyper-parameters. In dSprites, we followed standard practice \cite{ chen2018isolating} and quantified disentanglement in the training data. 
All networks were implemented with PyTorch 
and optimized with Adam \cite{kingma2014adam}. All statistical tests 
were based on paired Student's \textit{t}-tests.

\subsection{Qualitative benchmarks}
\subsubsection{Colored MNIST}
\label{sec:coloredMNIST}
We augmented the 
binary
MNIST data set \cite{lecun1998gradient}
by adding red, green and blue color to 3/4$^{th}$ of the white characters, resulting in 4 types of colors in the data set with an input size of 2352 (3*28*28).
This added a discrete nuisance factor to the  
inherent \textit{style} variations in the original data set. 
We focused on the ability of IBP-VAE 
to disentangle color and other style variations in comparison to VAE. 
Both the encoder and decoder consisted of two hidden layers of 500 units, each with ReLU activation. 
$\bm{\mu}, \bm{\sigma}^2$ and $\mathbf{d}$ in (\ref{eq:qDecomposition})
were further obtained with one hidden layer, with the truncation number $K$ 
set to $100$ and 
parameter $\alpha$ set to $30$ for the Beta distribution. For optimization, we used a learning rate of 1e-4.

Fig. \ref{fig:MNIST_disentanglement} gives examples of latent space traversal of the trained IBP-VAE and VAE. As shown, 
IBP-VAE disentangled semantically meaningful  
factors such as rotation (a), digit type (c), 
stroke width (e), and color (f).
In specific, 
IBP-VAE learned to 
encode the presence of font color by the activation of 
a specific latent unit: de-activation of this unit could  
independently remove the font color 
as demonstrated in Fig. \ref{fig:MNIST_disentanglement}(f). 
We refer to this as 
a \textit{triggering} unit in the rest of the paper.
In comparison, VAE 
was not able to 
disentangle color from either rotation (b) or digit type (d), 
nor was it able to 
extract the 
generative factor of
\textit{stroke width}.  

\subsubsection{3D Chairs}
The data set of 3D chairs  \cite{aubry2014seeing},  extensively considered for qualitative demonstration of disentangled representation learning, 
comprises of factors of variations such as rotation, width, and leg style of the chairs. 
Here, 
we compared the disentangling ability of $\beta$-IBP-VAE to  $\beta$-VAE using the same experimental setup and network architecture as 
\cite{higgins2016beta}.

Fig. \ref{fig:3D_chair} shows the results of
latent space traversal of $\beta$-IBP-VAE and $\beta$-VAE at $\beta=10$, the value at which we obtained the best results for $\beta$-VAE.  
Similar to what was shown in \cite{higgins2016beta}, 
$\beta$-VAE captured three factors of variation including azimuth, width, and leg style. 
In comparison,  $\beta$-IBP-VAE 
was able 
to disentangle the same three factors, along with an additional  generative factor: the height of the chair. 
Moreover, $\beta$-IBP-VAE seemed to have found a binary \textit{triggering} unit that swaps between two different leg styles
(Fig. \ref{fig:3D_chair} bottom panel).

\subsection{Quantitative benchmark}
We quantitatively evaluated 
IBP-VAE in two aspects. 

We first considered quantitative metrics 
recently proposed 
to measure disentanglement against available ground-truth factors of variation. 
In particular, 
we considered the metric of mutual information gap (MIG) \cite{chen2018isolating} that measures the normalized gap in mutual information between the top two dimensions in the latent vector that are most sensitive for each ground-truth factor. 
It is considered to 
addresses some of the limitations of previous metrics,
including that it is unbiased to hyperparameter settings and applicable to any latent distributions \cite{chen2018isolating}. 

The second analysis was inspired by the rate-distortion (RD) analysis introduced in \cite{alemi2018fixing}, which characterizes the competition between the first reconstruction term (distortion) and the second KL-divergence term (rate) in the ELBO objective (\ref{eq:VAEObj}). 
Here, 
we further narrowed down the RD analysis
to focus on the competition between the 
reconstruction and disentangling ability  
of the VAE. 
To do so, 
we singled out the 
total-correlation (TC) term from the 
rate term as 
shown in (\ref{eq:tc-decompo}),  
which measures the independence 
within the learned latent factors. 
We then contrasted it with the 
distortion term 
similar to the R-D analysis: 
we term this as TC-D analysis. 

Both quantitative analyses 
were carried out on 
the dSprites \cite{dsprites17} data set that consists of 737,280 synthetic images (64$\times$64) 
of 2D shapes with five known
generative factors: scale, rotation, x-position, y-position and shape. 
Beside VAE with a regular Gaussian prior, we also compared IBP-VAE with VAE with a complex prior in the form of VampPrior \cite{tomczak2017vae}. 
 For all models, 
we adopted the
CNN encoder-decoder architecture 
from  \cite{chen2018isolating} with a latent dimension 
of 10
all models 
(details in the Appendix \ref{supp:experiments}), 
and we varied the value of $\beta$ for the penalizing KL terms. 
A learning rate of 5e-4 and $\alpha$ value of 10 were used. Other hyper-parameters required for VampPrior were used as the standard implementation provided by \cite{tomczak2017vae}. 

\begin{table}[t]
\centering
\begin{tabular}[t]{|c|c|c|c|}
\hline
$\beta$ & \multicolumn{3}{c|}{MIG}  \\
& $\beta$-VAE & VampPrior & IBP-VAE\\
\hline
1 & 0.1890 &  0.1305 & \textbf{0.4174}  \\ 
5 & 0.4786 &   0.4848 & \textbf{0.5477} \\
10 & 0.4661 &  0.4676 & \textbf{0.485} \\
\hline 
\end{tabular} 
\caption{\small{The disentanglement score given by mutual information gap (MIG) from $\beta$-VAE, $\beta$-VampPrior and $\beta$-IBP-VAE.}}
\vspace{-0.3cm}
\label{tab:migAndFactor}
\end{table}

\begin{figure}[t]
\begin{center}
\includegraphics[scale=0.53]{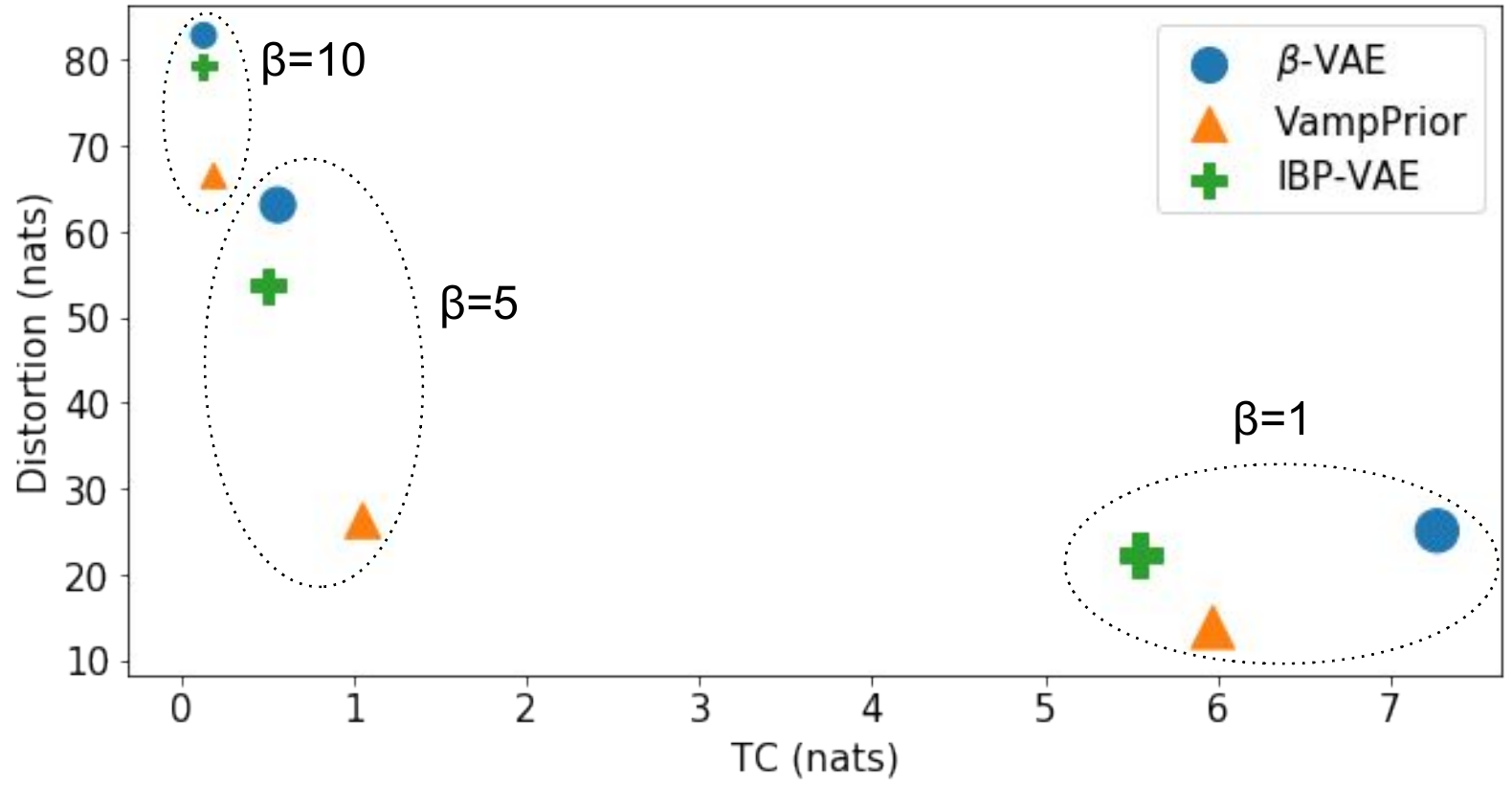}
\end{center}
\vspace{-.4cm}
\caption{\small{
TC-D analyses for $\beta$-VAE, -IBP-VAE, and -VampPrior on dSprites. The 
(TC, D) value obtained from each model is plotted for three different $\beta$ values. 
}}
\vspace{-.3cm}
\label{fig:RD}
\end{figure}


\subsubsection{MIG scores}
Table \ref{tab:migAndFactor} compares the MIG disentanglement scores of  $\beta$-VAE, $\beta$-VampPrior, and $\beta$-IBP-VAE at different values of $\beta$. 
Fig. \ref{fig:dSprites} 
visualizes the disentanglement performance of these models 
at $\beta = 5$, 
along with the best results adopted from \cite{chen2018isolating}. 
Each plot 
shows the relationship 
between the learned 
latent dimension (row) and the 
ground-truth factors (column): a successfully disentangled latent dimension 
should vary with only 
one ground-truth factor. 

As is evident 
both visually 
and quantitatively, 
across $\beta = 1, 5$ and $10$, 
$\beta$-IBP-VAE 
achieved better disentanglement 
than the other two models.
For instance, 
at $\beta = 5$ (Fig.~\ref{fig:dSprites}), 
$\beta$-IBP-VAE was able to clearly separate the rotation, scale, $x-$ and $y-$position. 
In comparison,
$\beta$-VAE heavily entangled rotation with position, 
while $\beta$-VampPrior captured the factor of rotation in two separate dimensions. 
Notably, 
at $\beta$=5, 
the MIG score reported by $\beta$-IBP-VAE surpassed the best result reported in  $\beta$-TCVAE \cite{chen2018isolating}, a state-of-the-art disentangling VAE that improves over $\beta$-VAE by penalizing only the TC term. 

\subsubsection{TC-D analysis}
In Fig. \ref{fig:RD}, we present the TC-D analysis for the three models considered. As we increased the $\beta$ value, 
the distortion (D) for all the models increased or, in other words, the ability of the model to reconstruct decreased. 
In the mean time, total correlation (TC) decreased, improving the independence among latent factors and hence helping in disentanglement.
In comparison to $\beta$-VAE with a regular Gaussian density, 
$\beta$-IBP-VAE was able to achieve lower distortion (better reconstruction) as well as lower or comparable TC (better or comparable independence) across all values of $\beta$. This verified our hypothesis that enabling richer yet independent posterior approximations 
was able to reduce the competition between the reconstructing and disentangling ability of VAE, allowing simultaneous improvement in both terms. 
VAE with the VampPrior, 
as expected, 
obtained the best reconstructions throughout all values of $\beta$ 
due to the use of a 
complex density. 
Without explicitly considering independence in the density, 
however, it resulted in decreased disentanglement compared to IBP-VAE, 
as measured by both the higher TC values (Fig.~\ref{fig:RD}) and the lower MIG values (Table~\ref{tab:migAndFactor}) across all values of $\beta$.




\begin{table}[t]
  \centering
  \begin{tabular}[t]{|c|c|c|c|}
    \hline
    Model & AC & AUC & MSE \\
    \hline
    CNN (AlexNet) & 82.59 &  0.75 & - \\ 
    c-VAE & 81.79 &  0.73 & 1521.05\\
    cIBP-VAE & \textbf{83.11} &  \textbf{0.79} & \textbf{1096.55}\\
    \hline 
  \end{tabular} 
  \caption{\small{Lesion classification accuray (AC), area under the ROC curve (AUC), and reconstruction mean square error (MSE) of cIBP-VAE, c-VAE, and baseline AlexNet.}\vspace{-.5cm}}
  \label{tab:skinResultAndRecon}
\end{table}
\begin{figure}[t]
\begin{center}
  \includegraphics[scale = 0.80]{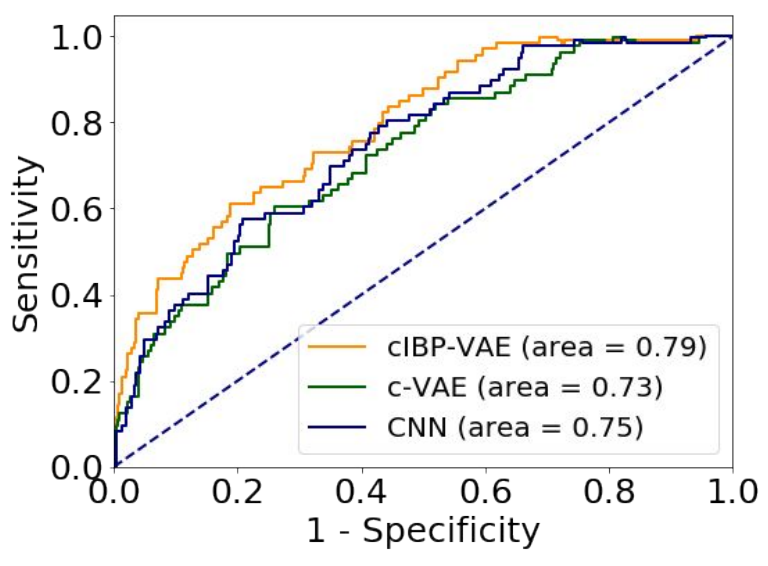}
  \end{center}
  \vspace{-.5cm}
\caption{\small{ROC curves of cIBP-VAE in comparison to alternative models for classification of melanoma and benign lesions}.\vspace{-.3cm}}
\label{fig:rocSkin}
\end{figure}


\begin{figure*}[t]
\begin{center}
  \includegraphics[width=\linewidth, height=8cm]{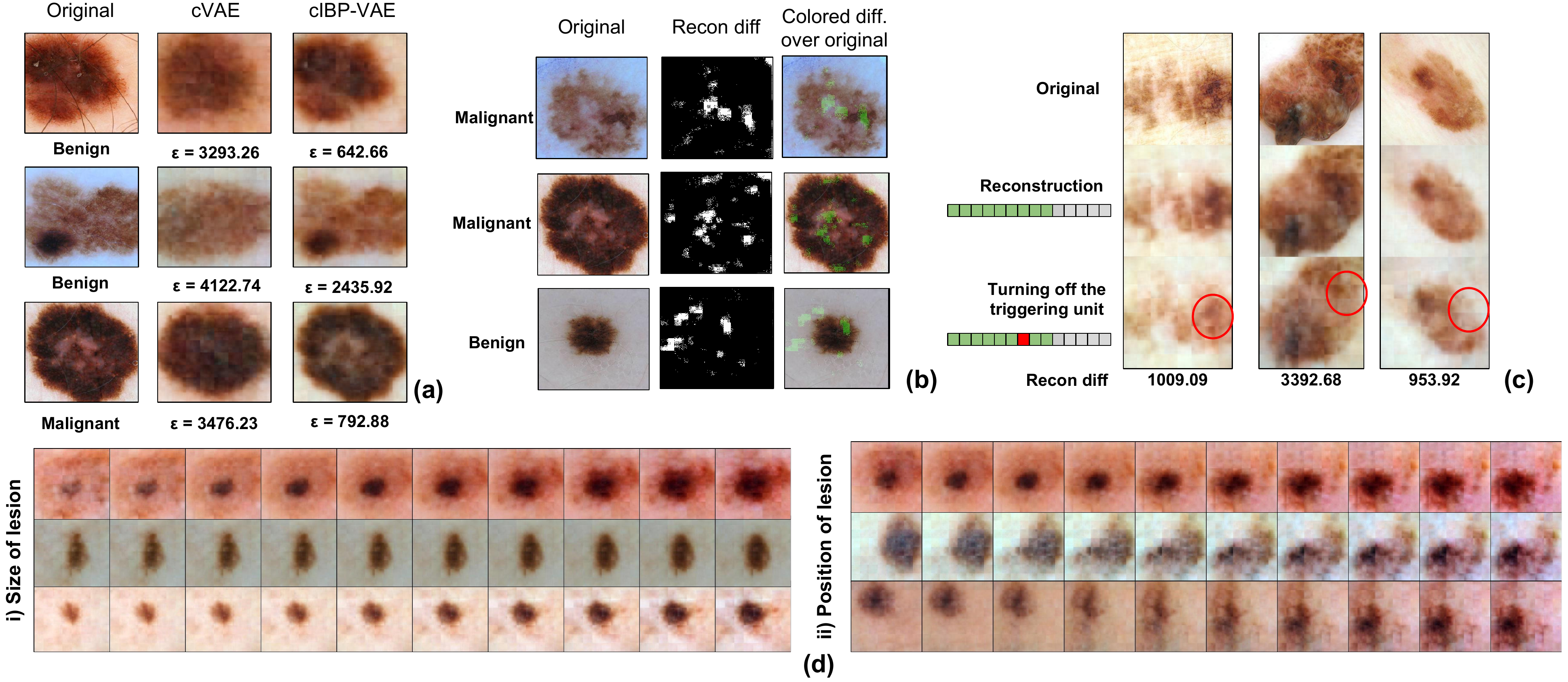}
  \end{center}
  \vspace{-.5cm}
\caption{\small{[Best viewed in color] (a) Reconstruction examples of cIBP-VAE and cVAE, along with MSE values. (b)  Column one: original images; column two: difference in reconstruction after switching the lesion label from malignant to benign (or vice versa); column three: overlay of reconstruction difference (green) with original images.  (c) Visual and quantitative reconstruction difference before (row two) and after (row three) de-activating
the \textit{triggering} unit. (d) Images generated by traversing a single latent unit over the [-5,5] range.}}
\label{fig:reconTriggerTraversal}
\end{figure*}

\subsection{Real-world clinical dataset}
\subsubsection{Skin lesion analysis}
ISIC
2016 \cite{gutman2016skin} 
is a public benchmark challenge data set 
consisting
of dermoscopic images of skin diseases released to support the development of melanoma
diagnosis algorithms. 
Here,
we considered the task of \textit{classification} of dermoscopic images into  melanoma (malignant) and benign categories. 
The challenge of this task 
lies in the need to 
extract subtle features 
relevant to melanoma detection, such as 
color and shape asymmetry \cite{soyer2004three}, from a large variety of lesion features. To be able to interpret semantically what factors did and did not contribute to the classification, therefore, is also important for the diagnosis decision.   

We used the 
given training and test with a size 900 and 379  images respectively. We further split a random 20\% of the training set for validation. Following pre-processing in \cite{li2018skin}, we  cropped the center portion of dermoscopic images and proportionally resized the cropped area to 256$\times$256. 
We used the AlexNet \cite{krizhevsky2014one}, pretrained on ImageNet dataset,
as the supervised baseline in this data set. 
The encoder in cIBP-VAE and c-VAE used the AlexNet to extract features, 
which 
were then factorized into $\mathbf{y}_{ns}$ and $\mathbf{y}_{ts}$ 
via two hidden layers on each branch. 
For $\mathbf{y}_{ns}$, both hidden layers used a size of 4096 and the truncation number $K$ was set to 50. For $\mathbf{y}_{ts}$, the two hidden layers used a size of 100 and 2 (representing class scores). For the decoder, we used the deep convolution architecture
(details in Appendix \ref{supp:experiments}). 
The value of $\zeta$ in (\ref{eq:finalObj}) was set to 5 with a \textit{warm-up} of 100  for 300 epochs \cite{bowman2015generating} and learning rate set to 1e-4.

\textbf{Task accuracy:} Table \ref{tab:skinResultAndRecon} summarizes the lesion classification performance of cIBP-VAE in comparision to the baseline discriminative AlexNet and c-VAE, using the two metrics recommended by ISIC for this task. The receiver operating characteristic (ROC) curves for all the models are also presented in Fig. \ref{fig:rocSkin}. 
As shown, 
while c-VAE decreased the task performance in comparison to AlexNet, 
cIBP-VAE significantly improved 
the lesion classification accuracy ($p <$ 0.04) and improved the ROC-AUC score from 0.75 (AlexNet) to 0.79. This suggests that unsupervised disentangling of nuisance factors could improve task accuracy if the nuisance factors are properly learned, 
and that VAE with a regular Gaussian density may have a limited ability to disentangle this data set given the complex factors of variations. 

\textbf{Uncovering and disentangling latent factors:}
To first compare the amount of factors of variations 
that could be captured by cIBP-VAE \textit{vs.} c-VAE, we compared the 
reconstruction accuracy 
of both models in test data. Table \ref{tab:skinResultAndRecon} (third column)
shows that the reconstruction error of cIBP-VAE was significantly lower ($p < 0.01$). 
Examples in Fig. \ref{fig:reconTriggerTraversal}(a) show that 
cIBP-VAE was particularly better at preserving the detailed color distribution in the skin lesion, which is known to be important for melanoma detection  \cite{soyer2004three}.   

To interpret the task-relevant representation learned by cIBP-VAE, 
we took  
the nuisance representation encoded by the cIBP-VAE from a test image  
and combined it with an  opposite image label for reconstruction. 
We expected the difference 
between the original and reconstructed images to explain what has contributed to the classification. 
Fig. \ref{fig:reconTriggerTraversal}(b) 
gives three such examples.  
Interestingly, 
after switching the label of a melanoma image, 
the reconstruction difference primarily focused on 
regions with asymmetry color or 
 atypical network within the lesion, 
 providing visual support on the subtle characteristics 
 that justified melanoma classification. 

Finally, 
to interpret the nuisance factors learned by cIBP-VAE, 
we analyzed images generated by traversing along continuous factors and 
de-activating 
binary factors. 
In Fig. \ref{fig:reconTriggerTraversal}(c), 
we show that 
cIBP-VAE has learned 
a triggering unit  
whose activation controls 
local lesion color,  
as highlighted by the red circle and the change in reconstruction error.  
In Fig. \ref{fig:reconTriggerTraversal}(d), 
we show images generated by traversing along two different latent dimensions learned by cIBP-VAE over a wide range of [-5, 5].
The results demonstrate that 
cIBP-VAE has discovered and disentangled 
factors such as the size and location of the lesion that are generally irrelevant to the task of melanoma detection. 




\subsubsection{Clinical 12-lead ECG}
\label{realData}
The ECG data set 
described in \cite{sapp2017real}
was collected 
during invasive electrical stimulation in the hearts 
of 39 post-infarction patients: 
on each patient, 
15-second 12-lead ECG recordings resulting 
from $19\pm11$ 
 different stimulation locations were collected. 
Following pre-processing  
in ~\cite{chen2017disentangling}, 
the data set consists of 16848 ECG beats 
(12 $\times$ 100, 12 = number of leads; 100 = temporal samples), 
each with a labeled 
site of electrical stimulation  
in the form of 
one of the ten anatomical segments of the left ventricle.
This data set was collected for the purpose of 
learning to localize the origin of ventricular activation from 12-lead ECG morphology, 
which can be useful for 
predicting the origin of abnormal rhythm 
in the heart and thus
guiding treatment. 

\begin{table}[t]
  \centering
  \begin{tabular}[t]{|c|c|c|}
    \hline
    Model & Seg. classification & Seg. classification \\
    & (in \%) & with artifacts (in \%)\\
    \hline
    CNN & 53.89 & 52. 44 \\ 
    c-VAE & 55.97 & 53.95 \\ 
    \textbf{cIBP-VAE} & \textbf{57.53} & \textbf{56.97} \\
    \hline 
  \end{tabular} 
  \caption{\small{Segment classification accuracy (with and without artifacts) of CNN, c-VAE, and cIBP-VAE on the  test set.}\vspace{-.5cm}}
  \label{tab:clinicalResult}
\end{table}

This task is challenged by 
significant inter-subject variations 
in a wide range of factors 
such as heart and thorax anatomy, 
heart pathological remodeling, 
and surface electrode positioning, 
all of which affect ECG morphology \cite{plonsey1969bioelectric}.  
Unlike visual disentangling in the last three data sets, 
these factors are also not  
directly visible on the data,  
but related to it through a complex 
physics-based process. 
To add a visual factor 
and to test the ability of cIBP-VAE to 
grow with the complexity of the data, 
we further
augmented this data set by an artifact 
(of size 10 for each lead) -- in the form of an artificial pacing stimulus -- to $\sim$50\% of randomly selected ECG data. 
The entire dataset was split into training, validation and test set, 
where no set shared data from the same patient.

The network architecture of IBP-VAE was identical to that used on colored-MNIST.
For 
$\mathbf{y}_{ts}$, a hidden layer of 10 units representing class scores was used. For nuisance factors $\mathbf{y}_{ns}$, 
batch-normalization 
was added after the encoded  representations 
with $\alpha$ = 20 for Beta distribution and the truncation number $K$ set to 50.
A learning rate of 1e-3 was used. For the weight hyperparameter $\zeta$ in equation (\ref{eq:finalObj}), values of 
$\{0.5, 0.8, 1\}$ were used to find the best model. 

We compared cIBP-VAE to: 1) 
a supervised CNN with three-layered convolution blocks (dropout, 2d convolution, batch normalization, ReLU, and max-pool layer) followed by two fully connected layers, 
and 2) c-VAE with the same parameters and architecture of cIBP-VAE. The design choice of the supervised CNN 
was inspired by  
\cite{yang2018localization}.

\begin{figure}[t]
\begin{center}
  \includegraphics[scale = 0.80]{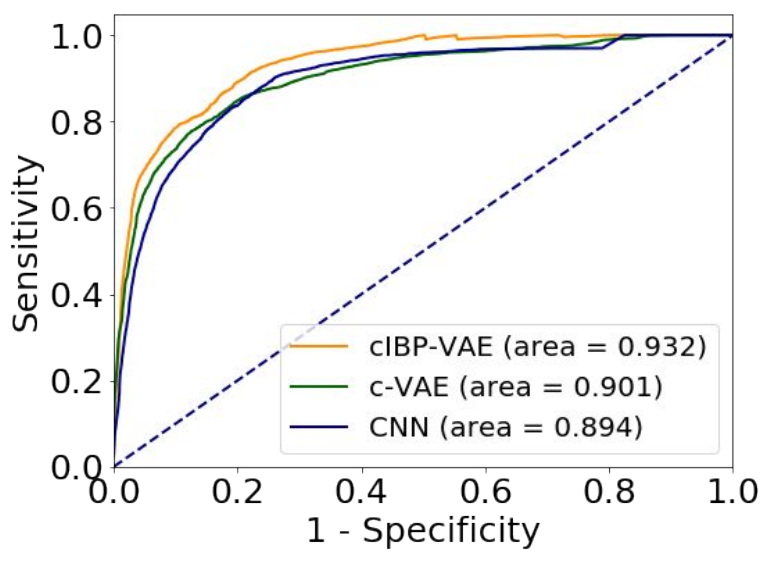}
  \end{center}
  \vspace{-.5cm}
\caption{\small{ROC curves of cIBP-VAE in comparison to alternative models on the clinical ECG data set}.\vspace{-.2cm}}
\label{fig:rocECG}
\end{figure}

\textbf{Task accuracy with increasing data complexity:} 
Table \ref{tab:clinicalResult} compares 
the classification accuracy on the test set 
obtained by the 
three models. 
The limited performance of CNN 
showed the significant challenge 
introduced by inter-subject variations 
on this data set. 
By adding unsupervised disentangling 
of nuisance factors, 
both c-VAE 
and cIBP-VAE achieved a higher classification,  
although cIBP-VAE 
significantly outperformed c-VAE 
either with or without the signal artifact 
($p < 0.03$).  
This improvement of performance 
is also summarized 
in the ROC curves in Fig. \ref{fig:rocECG},  
along with the value of 
the area under the macro-average ROC curve.  

It is also noteworthy that, 
while all models showed a decrease in classification accuracy when pacing artifacts were introduced to the data, 
cIBP-VAE exhibited the smallest margin of accuracy loss ($\downarrow 0.56\%$) in comparison to c-VAE ($\downarrow 2.02\%$) and CNN ($\downarrow 1.45\%$), 
further demonstrating the advantage of IBP-VAE 
to grow with the complexity of the factors of variations in the data. 

\begin{table}[t]
  \centering
  \begin{tabular}[t]{|c|c|c|c|c|}
    \hline
    model & all signal &  \multicolumn{3}{c|}{artifact segment}  \\\cline{3-5}
    &  & all & non-stimulus & stimulus\\
    \hline
    c-VAE & 2293.23 & 3.20 & 3.91 & 2.49 \\ 
    cIBP-VAE & 2273.65 & 0.45 & 0.19 & 0.72\\ 
    \hline
  \end{tabular} 
  \caption{Reconstruction errors of cIBP-VAE \textit{vs}. c-VAE 
  for the entire signals (column 2) and for the artifact segment only (columns 3-5). The latter is respectively calculated for all samples (\textit{all}), 
samples with no pacing artifact (\textit{non-stimulus}), 
and samples with pacing artifacts (\textit{stimulus}).\vspace{-.3cm}}
  \label{tab:reconResult}
\end{table}

\begin{figure}[tb]
\begin{center}
  \includegraphics[width=\linewidth, height=5cm]{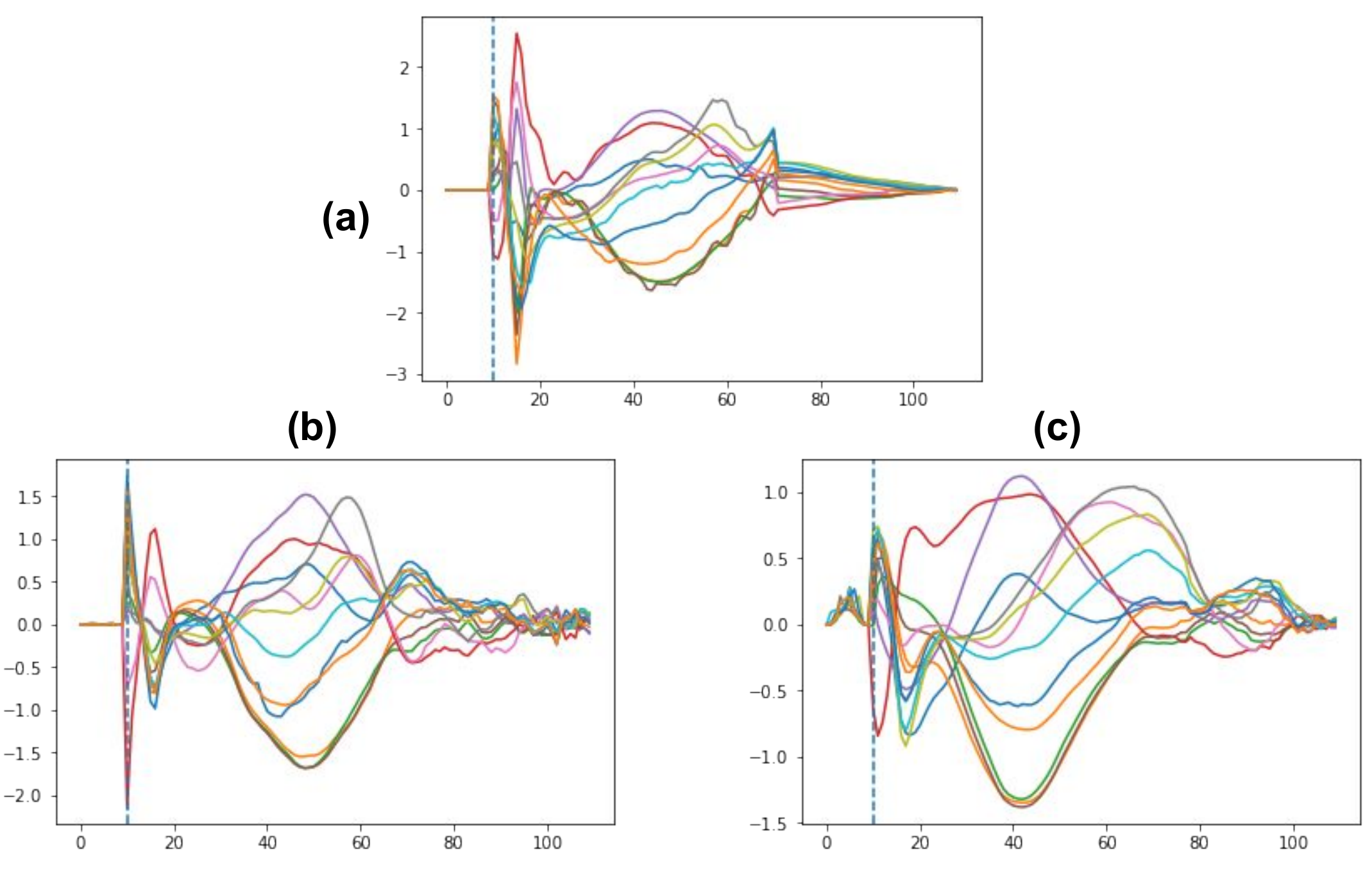}
  \end{center}
  \vspace{-.2cm}
\caption{\small{12-lead ECG traces where the pacing artifact is highlighted to the left side of the dotted line. \textbf{(a)} An original signal without the pacing artifact. \textbf{(b)} The reconstructed signal using cIBP-VAE. \textbf{(c)} The reconstructed signal using c-VAE.}\vspace{-.5cm}}
\label{fig:reconComparision}
\end{figure}

\textbf{Uncovering \& disentangling latent factors:} 
Because factors of inter-subject variations in the ECG data set cannot be labeled or directly visualized, 
here we 
focus on the ability of 
cIBP-VAE \textit{vs.} c-VAE in 
uncovering and disentangling 
the binary factor of pacing artifacts in the augmented data set.

\begin{figure*}[t]
\begin{center}
    \includegraphics[width=\linewidth]{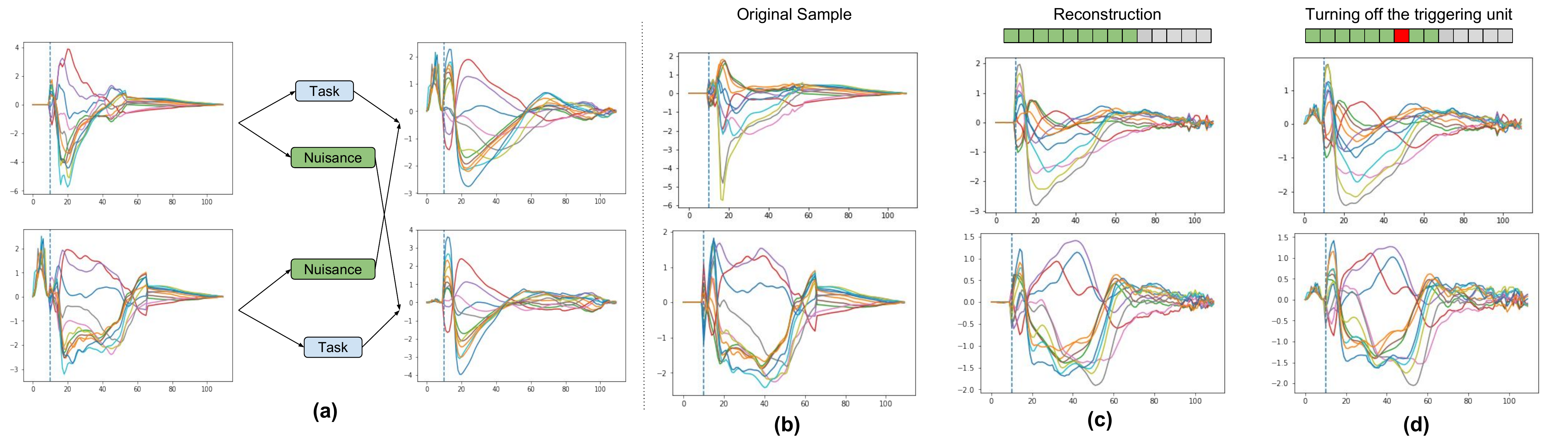}
  \end{center}
  \vspace{-.4cm}
\caption{\small{[Best viewed in color] \textbf{(a)}: Swapping between the task and nuisance representation from two samples (left) transferred the presence and absence of pacing artifact in the reconstructed signals (right).\textbf{(b)-(d)}:  Reconstructions before (c) and after (d) de-activation of the triggering unit, in comparison to original signals (b).  
}\vspace{-.2cm}}
\label{fig:trigger.}
\end{figure*}

As shown in Table \ref{tab:reconResult}, 
while cIBP-VAE and c-VAE 
showed a similar accuracy 
in reconstructing ECG signals, 
their accuracy in reconstructing 
the small artifact segment
differed significantly ($p<0.01$).  
Fig. \ref{fig:reconComparision} shows an example  
where 
cIBP-VAE was able to reconstruct the absence of a pacing artifact while c-VAE was not. 
This shows that cIBP-VAE 
was able to capture more generative factors, 
in a data set already riddled with 
a wide variety of factors of variations. 


To demonstrate the disentanglement of task and nuisance representations, 
we show in Fig. \ref{fig:trigger.}(a) that, when the encoded nuisance factors between a pair of signals were swapped,  
the absence and presence of the pacing artifacts were transferred as well.
Furthermore,
similar to previous data sets, 
cIBP-VAE has learned a 
\textit{triggering unit} 
to encode 
the absence or presence of the signal artifact in ECG data. 
Fig. \ref{fig:trigger.}(b)-(d) show two examples, 
where 
de-activation of this triggering unit 
added a pacing artifact 
to the reconstructed signal. 
This showed that cIBP-VAE 
was able to disentangle the specific nuisance factor of signal artifact, not only from the task  representation but also from  other nuisance factors.

\section{Conclusion and Future Work}
We presented a VAE model with a non-parametric independent latent factor model for unsupervised learning of disentangled representations. Departing from current focus on independence, we showed how an increased modeling capacity in the latent density
will improve the disentangling ability of VAE, especially as the complexity of the generative factors increases in the data. 
We further showed 
how unsupervised disentangling of nuisance factors could improve supervised extraction of task representations 
as well as 
facilitate interpretability of the learned representations. 
These were demonstrated through state-of-the-art qualitative and quantitative results on widely-used benchmark data sets, 
as well as 
improved performance over supervised deep networks on clinical data sets that have been little explored for the effectiveness of disentangled presentation learning. 
An immediate future work would include the extension of the current work to more benchmark data sets with a large number of variations and performing the quantitative comparison using additional disentanglement models and metrics \cite{locatello2018challenging}.

\appendices

\section{Model}
\subsection{Concrete distribution}
\label{sup:concrete}
During training of our presented IBP-VAE, we approximate Bernoulli random variables $Z$ with the \textit{Concrete} distribution \cite{maddison2016concrete} which has a convenient parametrization:
\begin{equation}
\begin{aligned}
\label{eq:gumbelSoftmaxSamples}
\tilde{z} = sigmoid \bigg(\frac{1}{\tau} \cdot (\log{(\pi_{k})} + g)\bigg)
\end{aligned}
\end{equation}
where $\tau$ is the temperature parameter, $g$ is the sample from the Gumbel(0,1) distribution which again has a convenient sampling strategy using Uniform(0,1).

\subsection{Kumaraswamy distribution}
\label{sup:kumaraswamy}
For the posterior involving Beta distribution, we approximate using the Kumaraswamy distribution as shown by \cite{nalisnick2016stick} where samples are drawn as:
\begin{equation}
\begin{aligned}
\label{eq:kumaraswamySamples}
\nu \sim (1-u^\frac{1}{b})^\frac{1}{a} 
\end{aligned}
\end{equation}
where u $\sim$ Uniform(0,1) and $a, b >$ 0. This distribution is equivalent to Beta distribution when $a$ = 1 or $b$ = 1 or both.



\section{Architecture Details}
\label{supp:experiments}
\begin{table}[!htbp]
  \centering
  \begin{tabular*}{\linewidth}[t]{c|c}
    \hline
Layer & Info (CNN Encoder for dSprites) \\
    \hline
    CNN$_{1}$ & 2dConv(1, 32, (4,4), 2, pad=1), ReLU \\
    \hline
    CNN$_{2/3/4}$ &2dConv(32, 32, (4,4), 2, pad=1), ReLU\\
    \hline
    FC$_{1}$ & Linear layer(256), ReLU \\
    \hline
    FC$_{2}$ & Linear layer(256), ReLU \\
    \hline 
    FC$_{3}$ & Linear layer(500), ReLU \\
    \hline
    $d(\cdot)$, $\mu(\cdot)$, diag($\sigma^{2}(\cdot)$) & 3 *  Linear layer(10) \\
    \hline
  \end{tabular*} 
  \label{tab:cnnArch1}
\end{table}

\begin{table}[!htbp]
  \centering
  \begin{tabular*}{\linewidth}[t]{c|c}
    \hline
Layer & Info (CNN Decoder for dSprites) \\
    \hline
    FC$_{1}$ & Linear layer(256), ReLU \\
    \hline
    FC$_{2}$ & Linear layer(4 *4 * 32), ReLU \\
    \hline
    CNN$_{1/2/3}$ & 2d Transpose Conv(32, 32, (4,4), 2, pad=1),  ReLU \\
    \hline
    CNN$_{4}$ & 2d Transpose Conv(32, 1, (4,4), 2, pad=1), Sigmoid \\
    \hline
  \end{tabular*} 
  \label{tab:cnnArch1}
\end{table}

\begin{table}[ht!]
  \begin{tabular*}{\linewidth}[t]{c|c}
    \hline
Layer & Info (CNN Decoder for Clinical Skin Analysis dataset ) \\
    \hline
    CNN$_{1}$ & 2d Transpose Conv(1, 64 * 8, (4,4), 1), \\ & batchNorm,  ReLU \\
    \hline
    CNN$_{2}$ & 2d Transpose Conv(64 * 8, 64 * 4, (4,4), 2, pad = 1), \\ & batchNorm,  ReLU\\
    \hline
    CNN$_{3}$ & 2d Transpose Conv(64 * 4, 64 * 2, (4,4), 4), \\ & batchNorm,  ReLU\\
    \hline
    CNN$_{4}$ & 2d Transpose Conv(64 * 2, 64, (4,4), 4), \\ & batchNorm,  ReLU\\
    \hline
    CNN$_{5}$ & 2d Transpose Conv(64, 3, (4,4), 2, pad = 1)\\
    \hline 
  \end{tabular*} 
  \label{tab:dcGANArch1}
  \vspace{-0.2cm}
\end{table}

\bibliographystyle{IEEEbib}
\bibliography{refs}

\end{document}